\documentclass{article}

\usepackage{arxiv}

\usepackage[utf8]{inputenc} 
\usepackage[T1]{fontenc}    
\usepackage{hyperref}       
\pdfoutput=1
\usepackage{url}            
\usepackage{booktabs}       
\usepackage{amsfonts}       
\usepackage{nicefrac}       
\usepackage{microtype}      
\usepackage{lipsum}
\usepackage{graphicx}
\graphicspath{ {./images/} }

\title{Small Object Detection by DETR via Information Augmentation and Adaptive Feature Fusion}

\author{
ji Huang \\
  School of Computing and Artificial Intelligence\\
  Southwest Jiaotong University\\
  Chengdu, China \\
  \texttt{HuangJi1019@my.swjtu,edu.cn} \\
   \And
 Hui Wang \\
  School of Electronics, Electrical Engineering \\ and Computer Science\\
  Queen's University Belfast\\
  Northern Ireland, UK \\
  \texttt{h.wang@qub.ac.uk} \\
}

\begin{document}
\maketitle
\begin{abstract}
Currently, the main challenge for small object detection algorithms is to ensure accuracy while pursuing real-time performance. The RT-DETR model performs well in real-time object detection, but performs poorly in small object detection accuracy. In order to compensate for the shortcomings of the RT-DETR model in small object detection, two key improvements are proposed in this study. Firstly, The RT-DETR utilises a Transformer that receives input solely from the final layer of Backbone features. This means that the Transformer's input only receives semantic information from the highest level of abstraction in the Deep Network, and ignores detailed information such as edges, texture or colour gradients that are critical to the location of small objects at lower levels of abstraction. Including only deep features can introduce additional background noise. This can have a negative impact on the accuracy of small object detection. To address this issue, we propose the fine-grained path augmentation method. This method helps to locate small objects more accurately by providing detailed information to the deep network. As a result, the input to the transformer contains both semantic and detailed information. Secondly, In RT-DETR, the decoder takes feature maps of different levels as input after concatenating them with equal weight. However, this operation is not effective in dealing with the complex relationship of multi-scale information captured by feature maps of different sizes. This limitation can lead to poor performance in detecting small objects. Therefore, we propose an adaptive feature fusion algorithm that assigns learnable parameters to each feature map from different levels. This allows the model to adaptively fuse feature maps from different levels and effectively integrate feature information from different scales. This enhances the model's ability to capture object features at different scales, thereby improving the accuracy of detecting small objects. Experimentally, our model outperforms the latest DETR model as the state of the art model on the Aquarium Object Detection Dataset.
\end{abstract}


\section{Introduction}

Object detection is a fundamental task in computer vision that involves identifying and localizing objects within an image or video. Object detection is used in many applications, including autonomous vehicles \cite{takumi2017multispectral}, surveillance \cite{joshi2012survey}, image understanding \cite{huang2011salient}, augmented reality \cite{ghasemi2022deep} and video retrieval \cite{amrani2020self}.
Small object detection aims to accurately locate and classify objects with an area of less than 32*32 pixels in an image or video \cite{tong2020recent}. Because small objects occupy limited information in the image and are easily affected by noise and deformation, small object detection has always been a difficult problem in the field of object detection.

In the field of small object detection, researchers have proposed various methods to address these challenges. Cascade-RCNN \cite{cai2018cascade} is a cascade detector that optimizes further at each stage by cascading multiple detectors. Gradually improves detection performance through multiple stages, particularly suitable for small objects. \cite{dubey2022improving} proposes SOF-DETR, a novel object detection model addressing the inductive bias of DETR. Utilizing normalized inductive bias through data fusion and introducing lazy-fusion of features, SOF-DETR enhances accuracy, particularly for small-sized objects. However, relatively high computational complexity, may not be suitable for real-time requirements. 

This problem triggered us to consider whether DETR (Detection Transformer) can be extended to real-time detection scenarios involving small objects to take full advantage of the end-to-end detector to improve the effectiveness of small object detection. To achieve this goal, we redesigned the model structure of RT-DETR to improve the performance of small object detection without a significant increase in computational effort.

Specifically, we found that in RT-DETR \cite{lv2023detrs}, a hybrid encoding of transformer and CNN is used, but only the last layer of features output from the Backbone is input into the Transformer for real-time object detection. This results in the input to the Transformer being the semantic information in the deep network and ignoring the detailed information in the shallow network that is important for small object localization, including edges, textures, or color gradients. In addition, in RT-DETR, feature maps of different levels are input to the decoder after a simple concatenate operation, which may lead to inadequate information fusion because feature maps of different sizes may capture semantic information at different levels and scales. A simple sum operation makes it difficult to handle such complex relationships of multi-scale information, which may lead to poor model performance in the detection of small or large objects.

Therefore, we propose a Fine-Grained Path Augmentation method to pass the low-layer detail information to the deep network, so that the input features of the Transformer have both semantic and detail information, which helps to locate small objects more accurately. At the same time, we introduce an adaptive feature fusion algorithm, which allows the model to adaptively fuse different levels of feature maps, effectively integrating feature information from different scales, so that the model can better capture the features of the object at various scales, and improve the accuracy of small object detection. This adaptive feature fusion algorithm has flexible parameters that can be adaptively adjusted according to different tasks and scenarios, enabling the model to better adapt to complex and changing visual environments.

\section{Related Work}
\subsection{Small Object Detection Techniques}
Small object detection has garnered consider such able attention in computer vision due to its importance in various applications, as surveillance, autonomous vehicles, and medical imaging. Cascade-RCNN \cite{cai2018cascade} introduces a cascaded detection strategy, progressively refining object localization and classification through multiple stages. This approach has proven effective in improving the robustness of small object detection. However, its higher computational demands may limit its applicability in scenarios requiring real-time processing. \cite{lim2021small} presents a object detection method that uses context to improve the accuracy of small object detection. The method uses additional features from different layers as context by connecting multiscale features, and introduces object detection with an attention mechanism that enables it to focus on the object in the image and include contextual information from the object layer While these methods have made significant strides in enhancing small object detection, several challenges persist. The sensitivity of current approaches to noise, occlusion, and computational demands poses limitations in real-world scenarios. Additionally, there is a need for lightweight models that balance accuracy and efficiency, especially for applications requiring real-time responsiveness. 
\subsection{Real-Time Object Detection Techniques}
Real-time object detection is a critical requirement for various applications, including robotics, autonomous vehicles, and video surveillance. The YOLO (You Only Look Once) series, including YOLOv2, YOLOv3, and YOLOv4, has been instrumental in pushing the boundaries of real-time object detection. These models adopt a one-stage architecture, enabling them to achieve impressive detection accuracy while maintaining high processing speeds. EfficientDet \cite{tan2020efficientdet} is designed to strike a balance between accuracy and computational efficiency. Leveraging a compound scaling method. Its ability to provide high accuracy with fewer computational resources makes it particularly appealing for resource-constrained applications. SSD \cite{liu2016ssd} is known for its capability to perform multi-scale object detection in real-time. By utilizing feature maps from different scales, SSD achieves a good trade-off between accuracy and speed. Its design emphasizes efficiency, making it suitable for applications where real-time processing is crucial. 

While recent advances have significantly improved the real-time object detection landscape, challenges persist. The demand for even faster processing speeds without compromising accuracy remains a key concern. Additionally, handling occlusion, complex scenes, and diverse object scales in real-time scenarios requires ongoing attention.
\section{Methodology}
\subsection{Overview of RT-DETR}
RT-DETR is a real-time end-to-end object detector that leverages Vision Transformers (ViT) for efficient processing of multiscale features, thereby delivering real-time performance while maintaining high accuracy. The backbone network of RT-DETR employs a CNN architecture, such as the popular ResNet series or Baidu's in-house developed HGNet. The encoder component of RT-DETR adopts an efficient hybrid encoder, addressing multiscale features through the decoupling of internal scale interactions and cross-scale fusion. This unique Vision Transformers architecture reduces computational costs, enabling real-time object detection. The decoder portion of RT-DETR utilizes a multi-layer Transformer decoder, allowing the flexibility to choose different numbers of decoder layers during inference, thus adaptively adjusting inference speed without the need for retraining.
\begin{figure} 
    \centering
    \includegraphics[width=0.9\linewidth]{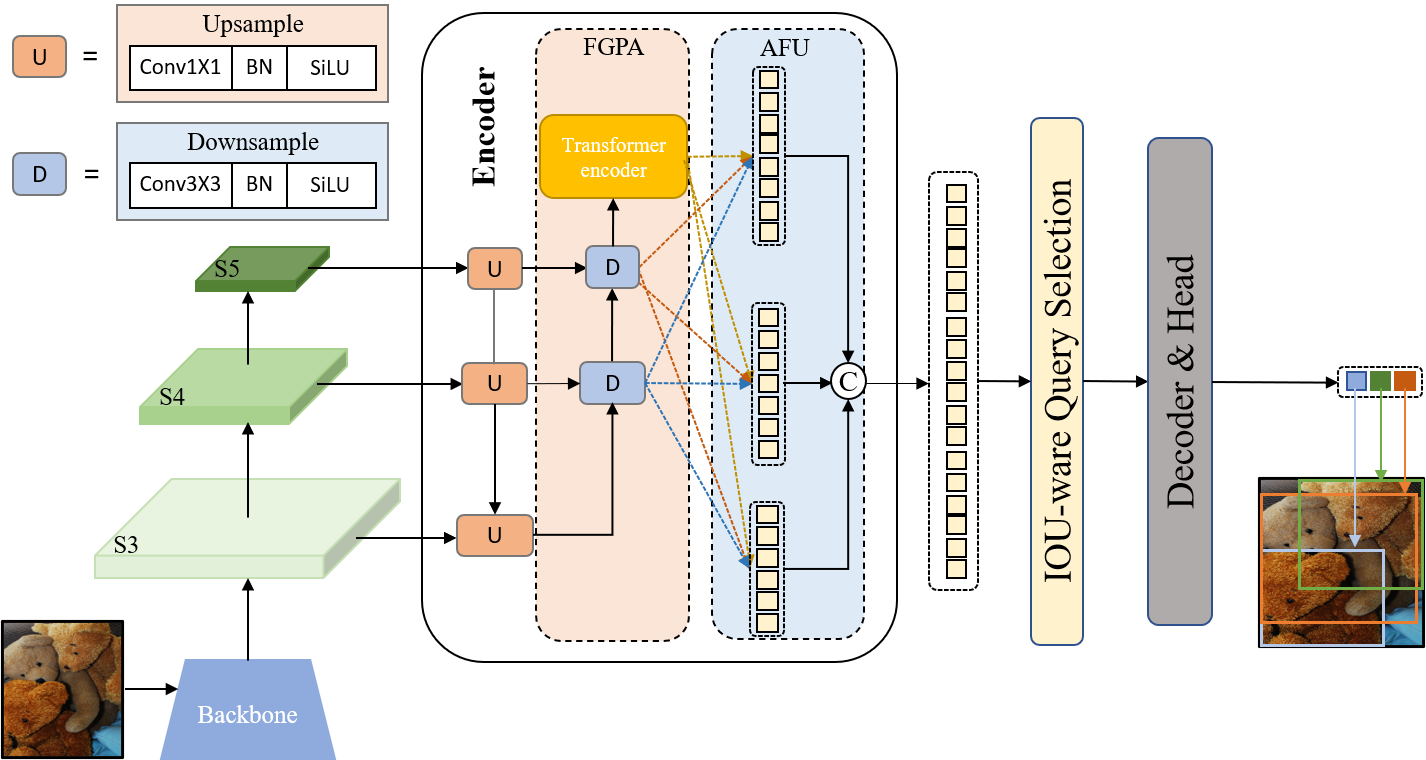}
    \caption{The overview of the model.FGPA means Fine-Grined Path Augmentation. AFU means Adoptively Feature Fusion.}
\end{figure}
\subsection{Fine-Grained Path Augmentation}
RT-DETR facilitates the detection and recognition of objects in an image by applying the self-attention mechanism to high-level features with richer semantic concepts to capture the links between conceptual entities in an image, which in turn helps subsequent modules detect and recognize objects in an image. However, RT-DETR considers the in-scale interaction of low-level features unnecessary due to the lack of explicit semantic concepts in low-level features and the risk of duplication and confusion in the interaction with high-level features.

Considering that the underlying feature maps of neural networks contain more local information and details that are crucial for processing complex images and small object image tasks. For this reason, we propose "fine-grained bootstrap network enhancement", which improves the feature information flow by introducing additional paths from the bottom layer to the top layer. The main principle is to introduce additional bottom-up paths on top-down paths to ensure that the features input to the Transformer include both the semantic information of the deep neural network and the local, detailed features of the lower neural network, thus improving the performance of the network on the small-object detection task.

In traditional deep neural networks, information may be lost or blurred between different layers. By introducing the underlying path, we can mitigate this information loss, which helps to retain more details and improve the perceptual ability of the model. Finally, the introduction of underlying paths helps to improve the flow of gradients through the network, making it easier to update the underlying weights, thus mitigating the problem of vanishing or exploding gradients and promoting more stable training.

\subsection{Adoptively Feature Fusion}
RT-DETR uses a feature fusion approach where feature maps from different levels or resolutions are directly stitched together. Although this approach is simple and intuitive, it is difficult for the model to distinguish which parts of the features are more important to the task and may introduce redundant information. In addition, the concatenate operation uses fixed weights to stitch the features together, which is not flexible enough to deal with multi-scale information or different tasks and cannot adapt to the needs of different scenarios or tasks for different scale features. In the small object detection task, simple concatenating easily ignores the modeling of local information, which is crucial for the accuracy and performance of small object detection.

For this reason, this paper proposes an adaptive feature fusion approach to better adapt the model to the small object detection task. Adaptive feature fusion refers to the fusion of feature maps from different levels or resolutions dynamically and flexibly in a deep learning model to improve the performance of the model on the task. This process adapts itself according to the input data or the needs of the task to better fit complex scenes and multi-scale information.
Specifically, we introduce a learned attentional weight matrix A with convolutional operations and activation functions to dynamically adjust the weights of the feature maps. The final fused feature Y is obtained by element-by-element multiplication, where the weights are determined by the attention weight matrix. The specific implementation process is as follows:
Let the input feature be \(X\in R^{C\times H \times W}\), where \(C\) is the number of channels, and H and W are the height and width of the feature map, respectively. Consider introducing a learned attentional weight \(A\in R^{H\times H}\), where the weight of each position indicates the contribution of the corresponding position in the feature fusion.
\begin{equation}
    A=\sigma(W_a \ast X)
\end{equation}

where \(\ast\) denotes the convolution operation, \(W_a\) denotes the learnable parameter with the shape \((k,k,C,1)\), \(k\) is the size of the convolution kernel, and \(C\) is the number of channels of the input features. \(\sigma\) is the activation function, and \(X\) is the input features.
The final fused features are:
\begin{equation}
    Y = \sum^H _{i=1} \sum^H _{j=1} A_{ij} \cdot X_{ij}
\end{equation}

where \(Y\in R^{C}\) is the final fused feature vector, \(A_{ij}\) is the value of position \((i,j)\) in the attention weight matrix, and \(X_{ij} \in R^C\) is the feature vector of the corresponding position.

This approach uses learnable weights that are updated through the training process of the network to ensure that the fused features better meet the requirements of the model for the task. This flexibility allows the model to adapt itself between different input scenarios and tasks, improving the generalization ability of the model.
\begin{table}
 \caption{Comparision with other models}
  \centering
  \begin{tabular}{lllllll}
    \toprule
     & \(AP^{val} \)    & \(AP^{val} \)    & \(AP^{val} \)  &\(AP^{val} \) & \(AP^{val} \)  &\(AP^{val} \)\\
    \midrule
    DETR  & 0.5	& 1.9	&0.3	&0.3	&0.5	&1    \\
    DINO  & 29.1 &52.5	&27.2	&10.7	&21.1	&40.8      \\
    RT-DETR    &29.1	&52.5	&27.2	&10.7	&21.1	&40.8  \\
    Ours &51.1	&80.9	&52.7	&22.3	&41.7	&63.4 \\
    \bottomrule
  \end{tabular}
  \label{tab:table}
\end{table}
\section{Experiments}
\paragraph{Dataset} The Aquarium Object Detection Dataset is collected by Brad Dwyer(Roboflow team) from two aquariums in the United States: The Henry Doorly Zoo in Omaha (October 16, 2020) and the National Aquarium in Baltimore (November 14, 2020). The dataset consists of 638 images splitted into train, test and validation data. The following classes are labeled: fish, jellyfish, penguins, sharks, puffins, stingrays, and starfish. Most images contain multiple bounding boxes. Since aquarium environments may contain a variety of water currents, reflections, occlusions, and other complex factors, this makes the dataset challenging for the object detection task .

\paragraph{Implementation Details} We use HGNetv2 \cite{lv2023detrs} pretrained on ImageNet \cite{russakovsky2015imagenet} with SSLD \cite{cui2021beyond} from paddleClas as our backbone. The structure of the transformer encoder and the structure of the decoder almost follow the RT-DETR\cite{lv2023detrs}. The training strategy and hyperparameters of the decoder almost follow the DINO \cite{zhang2022dino}.
\subsection{Comparision with SOTA}

In our experiments, we use a number of evaluation metrics to comprehensively assess the performance of the object detection model. These metrics include \(AP^{val}\)  (average accuracy on the validation set), \(AP^{50}\)(average accuracy at \(50\%\) IoU threshold), \(AP^{75}\) (average accuracy at 75\% IoU threshold), and accuracy for different object sizes, i.e., \(AP^S\), \(AP^M\), and \(AP^L\) represent the average accuracy for small, medium, and large objects, respectively.

A detailed analysis of the performance of each model is then provided:
The DETR model shows a moderate level of performance in terms of \(AP^{val}\) at 0.5. However, its performance at \(50\%\) and \(75\% \)of IoU is relatively low, at 1.9 and 0.3 respectively, and its performance on small and large objects is also poor, at 0.3 and 1 respectively.

The DINO model performs well on all metrics, especially on \(AP^{val}\), \(AP^50\) and \(AP^{75}\) with 29.1, 52.5 and 27.2 respectively, showing high performance on small, medium and large objects.

The RT-DETR model achieved significant performance on all metrics, especially on \(AP^{val}\), \(AP^{50}\) and \(AP^{75}\) with 50.7, 81.2 and 52.6 respectively.
Our proposed model performs well on all metrics, especially on \(AP^{val}\), \(AP^{50}\) and \(AP^{75}\), reaching 51.1, 80.9, and 50.7, respectively. High performance is also demonstrated on small, medium, and large objects, which is slightly improved compared to the RT-DETR model.

In our experiments, the detection of small objects becomes a key focus as it relates to a challenging aspect of the object detection task. First, the DETR model performs relatively poorly at detecting small objects, showing relatively low performance with an average accuracy \(AP^S\) of only 0.3 for small objects. In contrast, the DINO model performs well on small objects with a high mean accuracy, specifically its mean accuracy \(AP^S\) on small objects is 10.7, indicating strong performance on the small object detection task. In addition, the RT-DETR model has made significant progress in small object detection, with an average accuracy \(AP^S\) of 17.6 on small objects, showing relatively high performance. Finally, our proposed model also performs well in small object detection, with an average accuracy \(AP^S\) of 22.3 for small objects, which is slightly improved compared to the RT-DETR model. This set of experimental results highlights the superior performance of our proposed model in handling the small object detection task, which provides strong support for the accurate detection of small objects in practical application scenarios.
Overall, our model performs well on the \(AP^{val}\), \(AP^{70}\), \(AP^S\) ,and \(AP^L\) metrics, approaching or even surpassing the state of art model.

\subsection{Ablation experiments}
\begin{table}
 \caption{Validate the validity of each module}
  \centering
  \begin{tabular}{lllllll}
    \toprule
     & \(AP^{val} \)    & \(AP^{val} \)    & \(AP^{val} \)  &\(AP^{val} \) & \(AP^{val} \)  &\(AP^{val} \)\\
    \midrule
    RT-DETR  &50.7	&81.2	&52.6	&17.6	&42.2	&62.4   \\
    RT-DETR+FGGAN  &51.0	&79.8	&53.5	&21.6	&41.2	&62.8      \\
    RT-DETR+AFF   &50.8	&80.3	&50.5	&18.2	&41.1	&63.4  \\
    RT-DETR+AFF &51.1	&80.9	&50.7	&22.3	&41.7	&63.4 \\
    \bottomrule
  \end{tabular}
  \label{tab:table2}
\end{table}

In our small object detection ablation experiments, we improve the RT-DETR model by introducing two modules, namely Fine-Grained Guiding Network Augmentation (FGGAN) and Adoptive Feature Fusion (AFF). The performance of each model on different metrics is shown below:

For RT-DETR, the performance on \(AP^{val}\), \(AP^{50}\) and \(AP^{75}\)  is 50.7, 81.2 and 52.6 respectively, which is a BASELINE performance. For the small object (\(AP^S\)) the performance was 17.6, for the medium object (\(AP^M\) )  42.2 and for the large object (\(AP^L\)) 62.4.

On RT-DETR+FGGAN, \(AP^{val}\) was slightly better than RT-DETR with 51.0. On \(AP^{50}\) and \(AP^{75}\) it is 79.8 and 53.5 respectively, which shows some degradation in performance. On small objects (\(AP^S\)) the performance is 21.6 and on medium (\(AP^M\)) and large (\(AP^L\)) objects it is slightly reduced compared to RT-DETR.

On RT-DETR+AFF, performance is slightly better than RT-DETR on \(AP^val\) at 50.8.
Relatively stable performance on \(AP^50\) and \(AP^75\)  with 80.3 and 50.5 respectively. Performance was 18.2 on the small object (\(AP^S\)), 41.1 on the medium object (\(AP^M\)) and 63.4 on the large object \(AP^L\).

In RT-DETR+all, slightly improved compared to RT-DETR on \(AP^{val}\) with 51.1. Relatively stable performance on \(AP^{50}\)  and \(AP^{75}\) with 80.9 and 50.7 respectively. There was a slight improvement over RT-DETR on the small object (\(AP^S\)) with 22.3, on the medium object (\(AP^M\)) with 41.7 and on the large object (\(AP^L\)) it remained unchanged at 63.4.

\section{Conclusion }
In this study, we proposed two key improvements to address the challenges of balancing the accuracy and real-time performance of small object detection algorithms. By introducing underlying data augmentation and adaptive feature fusion algorithms, we enable the model to better capture detailed information and multi-scale features of small objects, thus improving the accuracy of small object detection. These improvements are simple and easy to implement, and experiments on the aquarium object detection dataset demonstrate that our model outperforms the latest DETR model, demonstrating its superior performance in practical applications. This provides strong support for more accurate and practical object detection models.

\bibliographystyle{unsrt}  
\bibliography{main}  

\end{document}